\documentclass[review]{elsarticle}

\usepackage{lineno,hyperref}

\usepackage{graphicx}
\usepackage{caption} 
\usepackage{subcaption}

\usepackage{algorithm}
\usepackage{algorithmic}

\modulolinenumbers[1]

\usepackage{url}

\journal{Journal of Artificial Intelligence}

\begin{document}

\begin{frontmatter}

\title{The Difficulty of Novelty Detection in Open-World Physical Domains: An Application to Angry Birds}

\author[inst1]{Vimukthini Pinto}
\cortext[mycorrespondingauthor]{Corresponding author}
\ead{vimukthini.inguruwattage@anu.edu.au}
\affiliation[inst1]{organization={School of Computing},
            addressline={The Australian National University}, 
            city={Canberra},
            country={Australia}}

\author[inst1]{Cheng Xue}
\author[inst1]{Chathura Gamage}
\author[inst2]{Matthew Stephenson}
\affiliation[inst2]{organization={College of Science and Engineering},
            addressline={Flinders University}, 
            city={Adelaide},
            country={Australia}}

\author[inst1]{Jochen Renz}




\begin{abstract}

Detecting and responding to novel situations in open-world environments is a key capability of human cognition and is a persistent problem for AI systems. In an open-world, novelties can appear in many different forms and may be easy or hard to detect. Therefore, to accurately evaluate the novelty detection capability of AI systems, it is necessary to investigate how difficult it may be to detect different types of novelty. In this paper, we propose a qualitative physics-based method to quantify the difficulty of novelty detection focusing on open-world physical domains. We apply our method in the popular physics simulation game Angry Birds, and conduct a user study across different novelties to validate our method. Results indicate that our calculated detection difficulties are in line with those of human users.
\end{abstract}


\end{frontmatter}

\section{Introduction}
With the increasing reliance on autonomous systems such as self-driving cars\cite{self-driving_cars2}, underwater exploration vehicles\cite{underwater_robot}, and planetary robots\cite{planetary_robots}, detection and adaptation to novel situations have become important capabilities for such AI systems. For example, if a self-driving car is not trained on slippery roads, the car may fail to detect that the friction is reduced and adjust the speed accordingly. Open-world learning is an emerging research area that attempts to address the challenge of detecting and adapting to novel situations \cite{Langley2020,Muhammad2021,PengX2021, Lee2021, Raj2021}. 
Open-world learning research requires adequate evaluation protocols to capture the performance of agents under the two tasks: detection and adaptation \cite{Senator2019}. This paper focuses on creating a difficulty measure for novelty detection to aid the evaluation of novelty detection by disentangling agents' performance from the intrinsic difficulty of novelties.

The novelties we encounter in an open world can take various forms \cite{Langley2020,Boult2021}. In this paper, we focus on \textit{structural transformation}, a very common type of real-world novelty where an unknown object is encountered or a previously known object changes one or more of its properties \cite{Langley2020}. For example, this could be a new vehicle type on the road, a new type of product in the supermarket with new packaging, a previously empty box filled with goods, or an abnormally heavy ball in a billiards game.
As these examples suggest, only some of the novelties can be identified from appearance. Novel objects with different appearances can be detected by observing the change in colour, shape, or size. Quantifying the difficulty of detecting them can be addressed with the use of concepts presented in colour science \cite{Giesel2010,Olkkonen2008} and research conducted on object shapes and sizes \cite{Perner2018,Zhao2005}. However, the difficulty of detecting novel objects with the same appearance but different physical parameters (e.g., mass, friction, bounciness) is not addressed so far. It is also not straightforward as one needs to interact with the objects and observe changes in their movements. Moreover, the detectability of such novelty depends on several factors, such as the physical parameter that is changed, or the number and arrangement of novel objects within the environment.

This paper presents a qualitative-physics based method to quantify the difficulty of detecting novel objects with the same appearance but altered physical parameters (compared to previously seen versions of these objects). 
The proposed method aids a thorough evaluation by disentangling agents' performance from the difficulty of detecting the novelty. 
For example, if the novelty cannot be identified from the appearance and occurs in an object that is not reachable to interact with, then the novelty cannot be detected. Therefore, the difficulty of novelty detection should be considered before making conclusions on the detection ability of an agent.  
The method we propose is agent independent and enables us to evaluate an agent's performance (both detection performance and task performance) at different levels of difficulty (that can be categorized as easy, medium, and hard). We apply our method to the popular physics simulation game Angry Birds, as it closely resembles real-world physics and provides an ideal platform to introduce novelty \cite{noveltygenerator, Xue2022}. We then conduct a user study experiment with human participants to verify that the calculated novelty detection difficulty values are in line with those of humans.

The rest of this paper is structured as follows. We start by providing the background and related work to our study, followed by the formulation of our novelty detection difficulty measure. We then present the application of our difficulty measure in Angry Birds. Next, we describe an experimental evaluation with human participants in Angry Birds to validate the proposed difficulty measure. Finally, we conclude the paper with possible improvements and future directions of this study.

\section{Background and Related Work}

This section presents the notion of difficulty and the concept of novelty in the context of physical worlds and AI. We also discuss the related work in physics simulation games, qualitative physics which is used in developing the difficulty measure, and a brief description of our experimental domain - Angry Birds. 

\subsection{Difficulty}

Assessing difficulty is a popular research area in neuroscience where researchers are interested in quantifying the difficulty of tasks or decisions \cite{Franco2018,Gilbert2012}. 
Measuring difficulty is also a main topic of discussion when measuring the intelligence of artificial systems \cite{MeasuringIntelligence1,MeasuringIntelligence2,Fernando2020,Fernando2019}. 
It is also a widely studied topic in the gaming industry to make games more interesting to players \cite{Togelius2008,Aponte2015,Mohammad2018,Dziedzic2018,Roohi2020}. The flow-theory, one of the most prevalent models in the game design literature, suggests that the games should not be too easy or too difficult to maintain player enthusiasm \cite{Takatalo2010}. 

Considering the difficulty of detection, researchers have studied this in areas such as emotion detection \cite{Laubert2019}, phishing message detection \cite{Steves2019}, and missing content detection \cite{YomTov2005}. However, to the best of our knowledge, the difficulty of novelty detection in physical domains has not been studied so far and quantifying difficulty becomes important when evaluating the detection capabilities of agents. 

\begin{figure}[t]
\centering
\begin{subfigure}{.433\textwidth}
  \centering
  \includegraphics[width=0.95\linewidth]{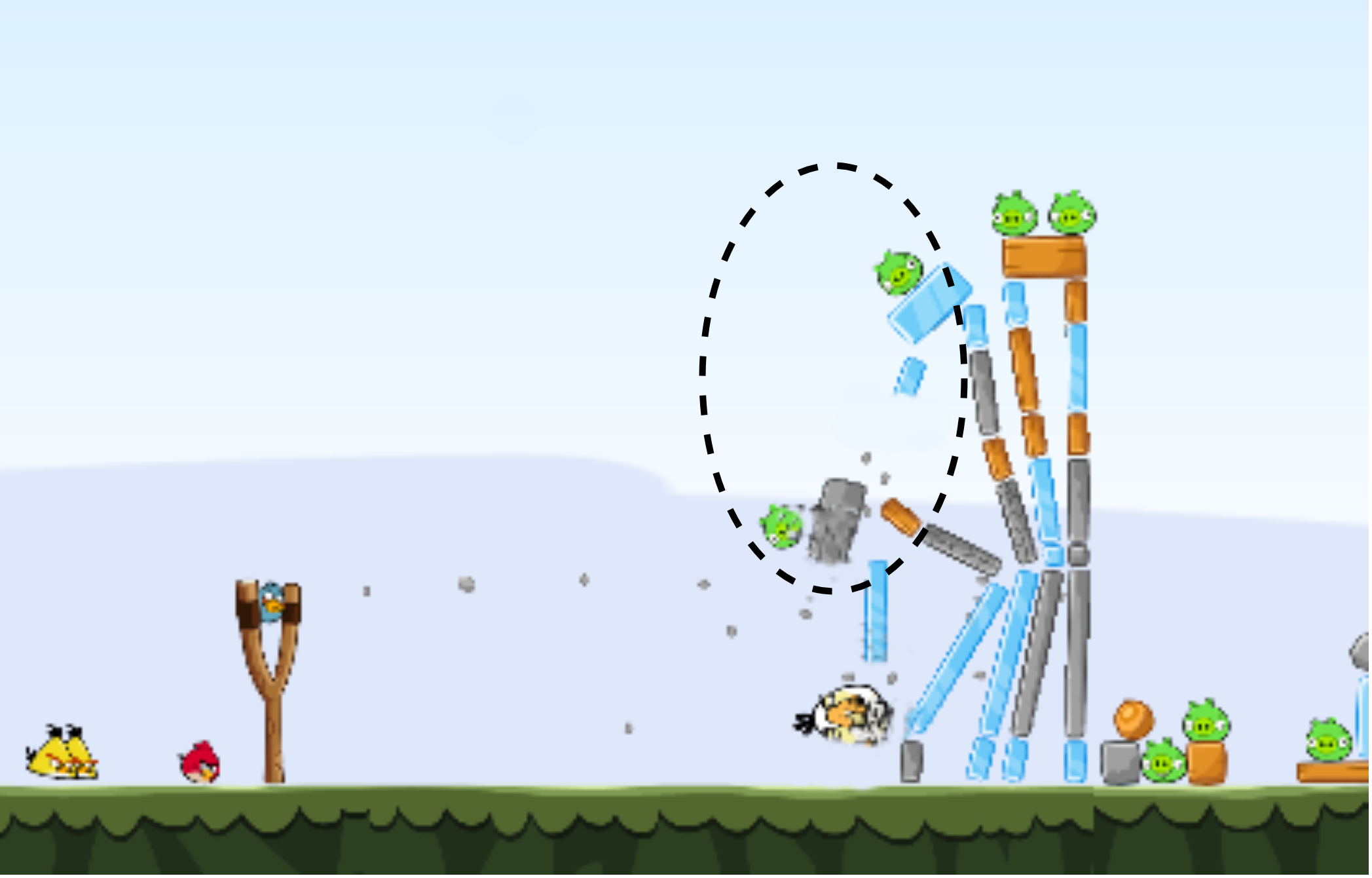}  
  \caption{}
  \label{fig1:sub-1}
\end{subfigure}
\begin{subfigure}{.433\textwidth}
  \centering
  \includegraphics[width=0.95\linewidth]{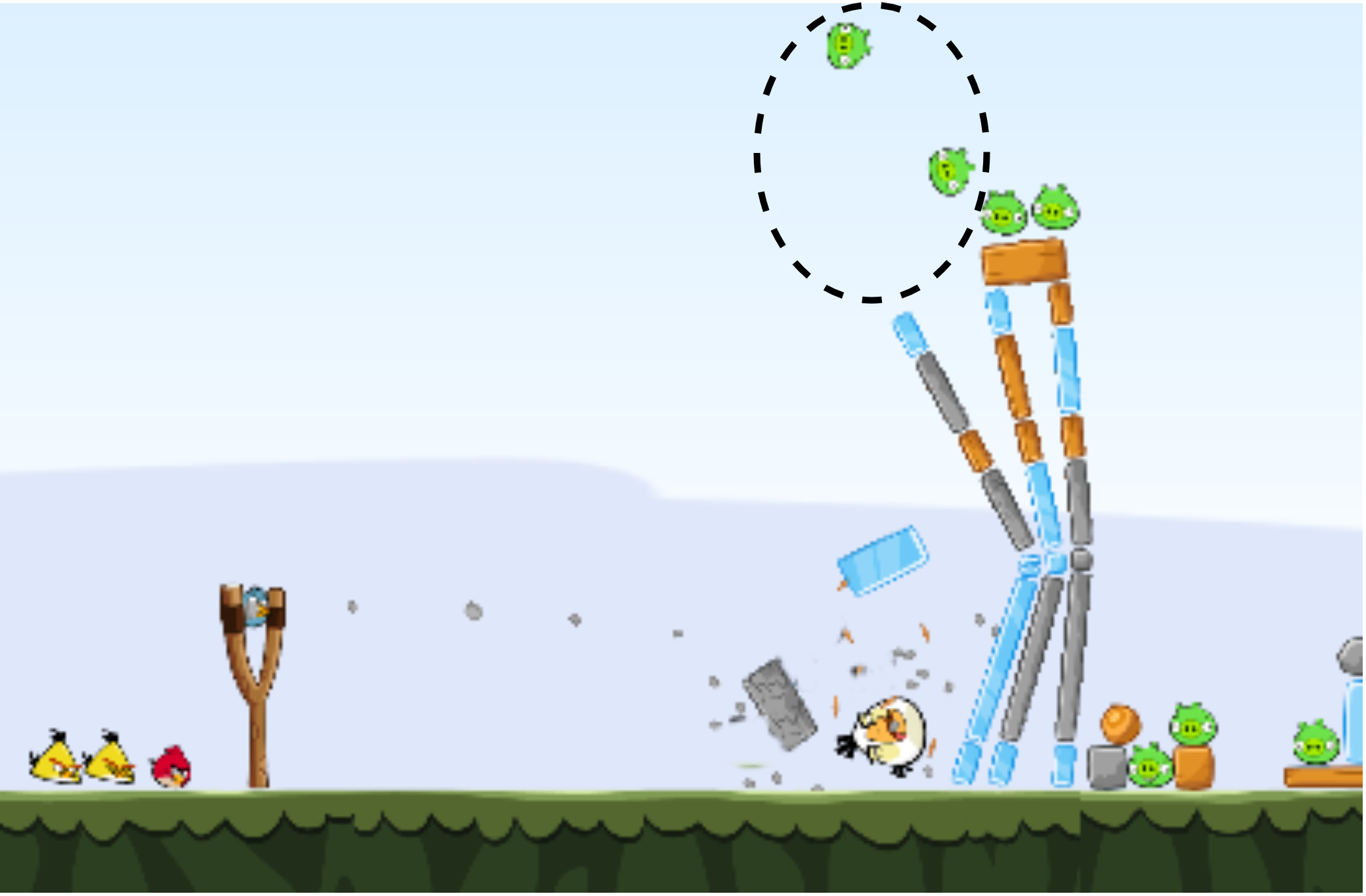}  
  \caption{}
  \label{fig1:sub-2}
\end{subfigure}

\begin{subfigure}{.2855\textwidth}
  \centering
  \includegraphics[width=0.95\linewidth]{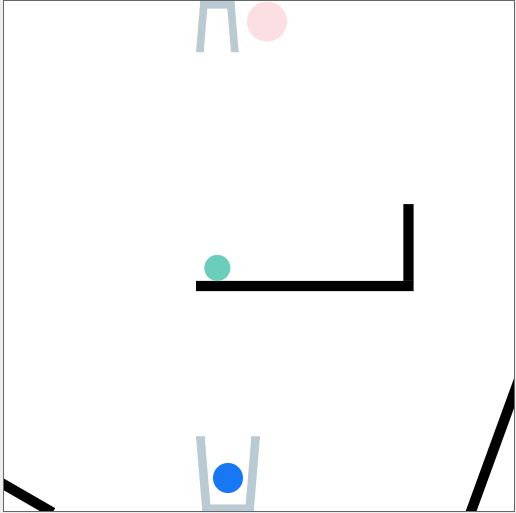}  
  \caption{}
  \label{fig1:sub-3}
\end{subfigure}
\begin{subfigure}{.2855\textwidth}
  \centering
  \includegraphics[width=0.95\linewidth]{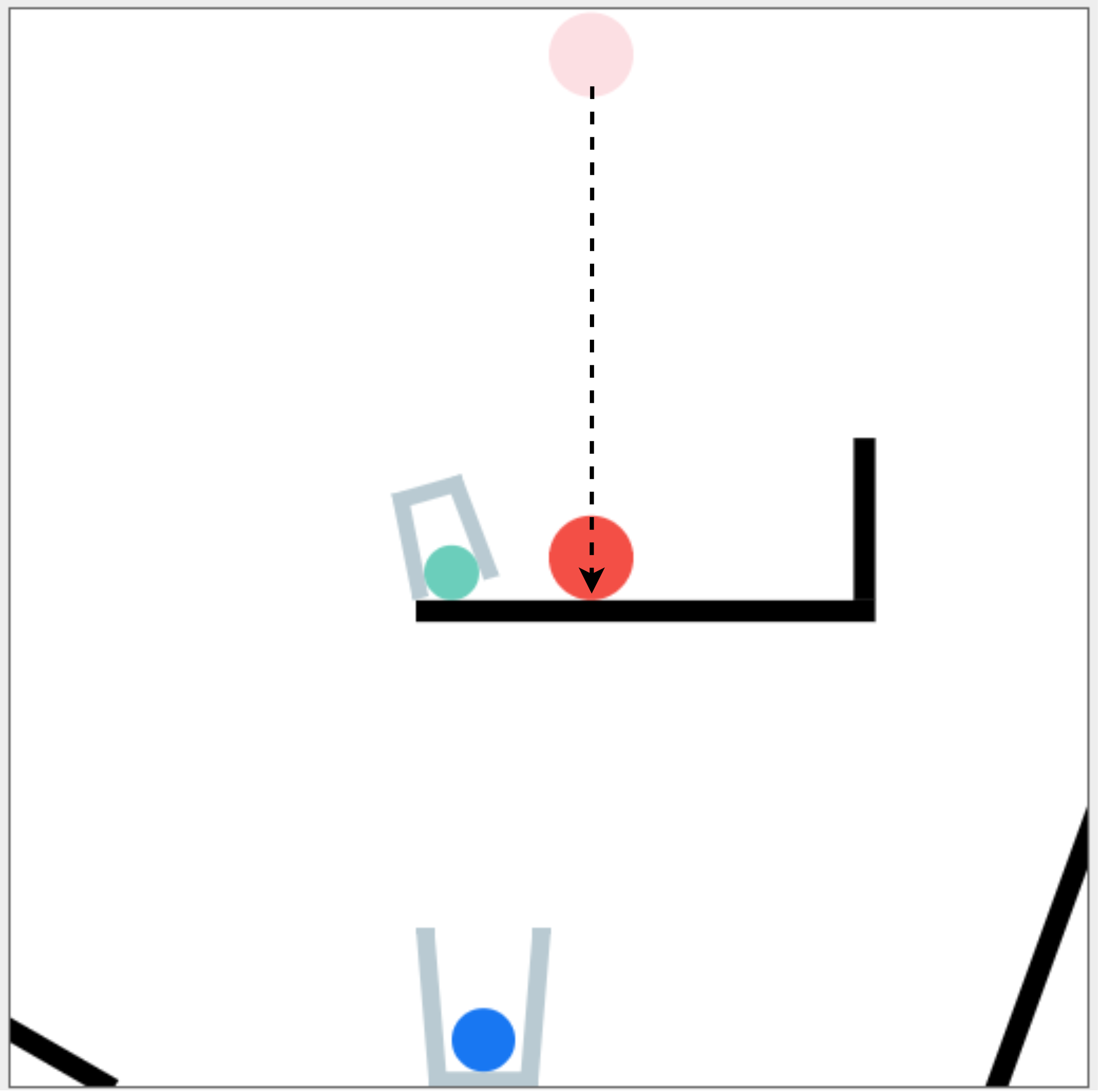}  
  \caption{}
  \label{fig1:sub-4}
\end{subfigure}
\begin{subfigure}{.2855\textwidth}
  \centering
  \includegraphics[width=0.95\linewidth]{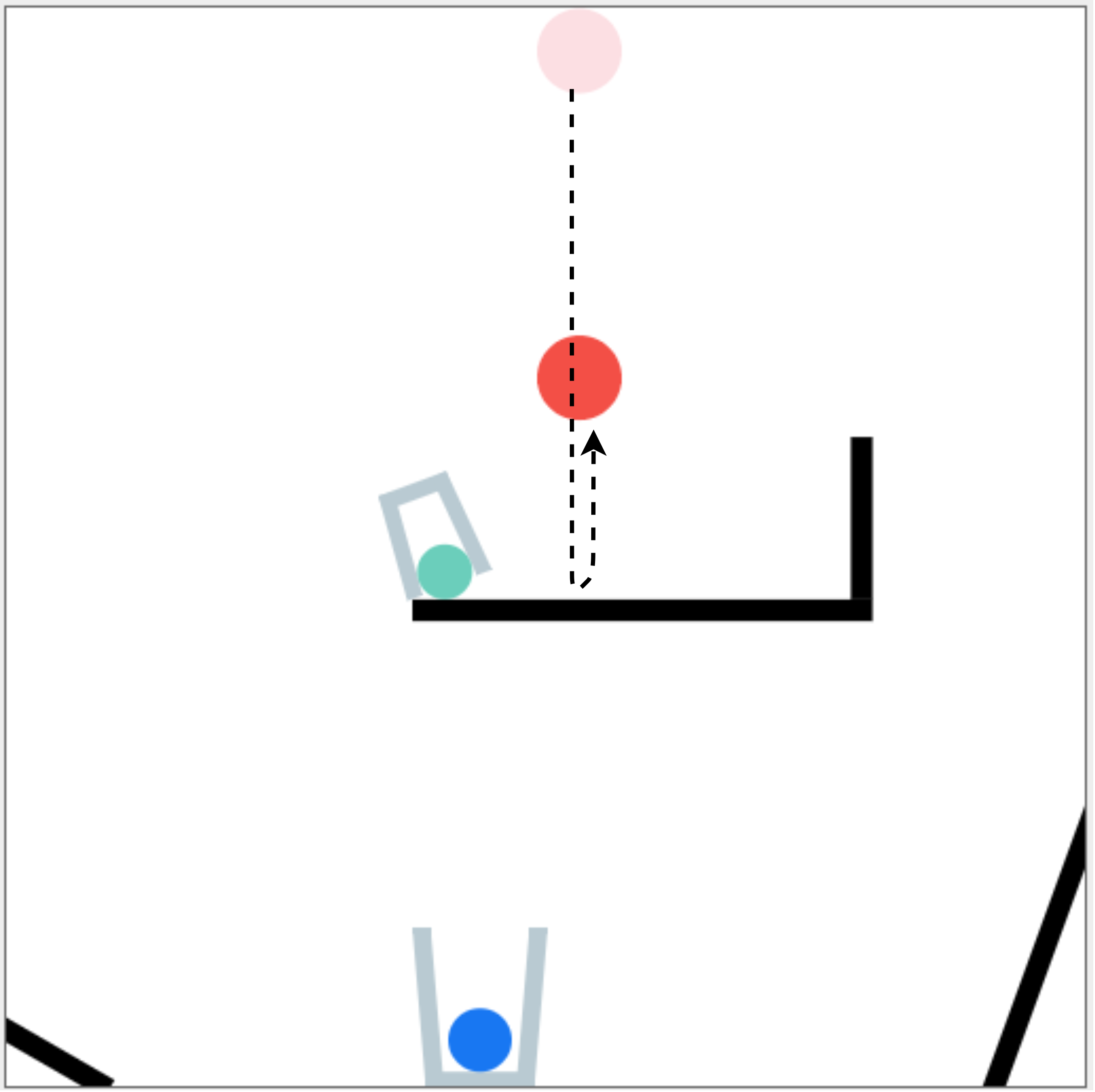}  
  \caption{}
  \label{fig1:sub-5}
\end{subfigure}
\caption{(a) and (b) are from Angry Birds domain (details on Angry Birds is available in Section \ref{Section Experimental Domain}). (a) has original parameters whereas (b) has an increased \textit{bounciness} parameter for pigs. The two figures show the difference in pig's movement for the same shot in the original (a) and increased value for \textit{bounciness} (b) of pigs. 
(c), (d), and (e) are from the PHYRE domain. (c) is the initial state of the game, (d) and (e) show the change in the height reached (after 188 frames) by the red ball with original parameters (d) and after (e) increasing the \textit{bounciness} parameter of the red ball. Details on PHYRE is available in Section \ref{Section Applying to PHYRE}.
}
\label{fig1}
\end{figure}

\subsection{Novelty} 

In the context of AI, novelty is described as situations that violate implicit or explicit assumptions about the agents, the environment, or their interactions \cite{SAILON}. 
A similar idea of novelty is presented by Muhammad et al. \cite{Muhammad2021}, where novelty is explained from an agent's perspective, i.e., when an agent encounters an entity, if it cannot recall the entity from prior experience, or cannot infer the entity through cognition, the encountered entity is considered novel for the agent.
Boult et al. \cite{Boult2021} formalizes a theory of novelty for open-world environments and Langley \cite{Langley2020} explains different types of environmental transformations that can be considered as novelty.
Following these ideas, the novelties we consider in this paper occur as a result of changed physical parameters of objects. It could be the mass, friction, elasticity, brittleness, etc. These novelties do not change the appearance of the object but cause it to behave differently after an interaction. For example, in the real-world, a novelty could be a new tennis ball with higher bounciness than the balls encountered before, a previously empty bottle now filled with water, or a box of goods with less weight due to a manufacturing defect. Figure \ref{fig1} shows differences in the observed movements of objects after physical parameters have been changed in two simulated physics environments: a research clone of Angry Birds \cite{Ferreira2014} and PHYRE \cite{Bakhtin2019}. 

\subsection{Physics Simulation Games}

A physics simulation game (PSG) is a video game where the game world simulates real-world physics and offers simplified environments for developing and testing AI agents \cite{Renz2017}. Game environments that require physical reasoning such as PHYRE \cite{Bakhtin2019}, Virtual Tools \cite{Allen2020}, OGRE \cite{AllenBakhtin2020}, IntPhys 2019 \cite{Riochet2020} and Physion \cite{Bear2021} have been developed due to the recent recognition of the importance of physical reasoning in AI. Angry Birds has also been a popular PSG for AI agents, with a long-running AIBirds competition being held every year since 2014 as part of the IJCAI conference \cite{AIBirds,Renzet2015}. 

PSGs are ideal platforms to introduce real-world novelties and to develop capabilities in AI to detect such changes. Boult et al.~\cite{Boult2021} have explained novelties that appear in the physics game CartPole \cite{CartPole}. The AIBirds competition \cite{AIBirds} has also introduced a track for agents to detect and adapt to novelty \cite{AIBirdsNT, Xue2022}.

\subsection{Qualitative Physics}

As discussed previously, novelties based on physics parameters are not detectable from appearance alone. Therefore, one needs to interact with the objects and observe any difference in their expected movements. Humans are often unaware of the exact physical parameters such as density, friction, and mass distribution of objects and do not need to solve complex differential equations when reasoning about their movements, instead relying on spatial intelligence \cite{Walega2016}. 


To analyze object movements, a qualitative physics approach was proposed by Zhang and Renz ~\cite{Peng2014}, which approximates structural stability based on the extended rectangular algebra (ERA). ERA comprises 27 interval relations based on the approximated centre of mass of the object and offers more flexibility than the original 13 interval algebra relations \cite{Allen1983}. Ge et al.~\cite{Gary2016} point out that ERA is more suitable to approximate the stability of a single object rather than a structure and extends the use of ERA by proposing two qualitative stability analysis algorithms: \textit{local stability} and \textit{global stability}. A similar algorithm, \textit{vertical impact} is proposed by Walega et al.~\cite{Walega2016}, which combines the concepts of {\it local stability} and {\it global stability} into one algorithm. Walega et al. also introduced the algorithm \textit{horizontal impact}, to provide a heuristic value to the interaction based on force propagation. 
Our difficulty measure also uses the algorithm \textit{vertical impact} \cite{Walega2016} along with new reasoning components which reason about the nature of the object movements that are necessary to detect novelty. 

\subsection{Experimental Domain} \label{Section Experimental Domain}

Our experimental domain, Angry Birds is a PSG where the player shoots birds at pigs from a slingshot. These pigs are often protected by different physical structures that are made up of three types of materials: wood, ice, and stone. 
There are also static platforms, which are indestructible objects that are not affected by forces.
The task of an agent that attempts to detect novelty is to identify if anything has changed from the normal game environment by shooting at game objects. 
As the original Angry Birds game by Rovio Entertainment \cite{Rovio} is not open source, we conduct our experiments using a research clone of the game \cite{Ferreira2014}. One example of novelty in Angry Birds is displayed in figures \ref{fig1:sub-1} and \ref{fig1:sub-2}. 
As Figure \ref{fig1:sub-1} shows, the agent who is familiar with the normal game environment expects the pigs to fall down without bouncing up after an interaction. However, when the bounciness parameter is increased, the agent observes a change in the pigs' movement as shown in Figure \ref{fig1:sub-2}.

We selected Angry Birds as our experimental domain due to three reasons. First, solving an Angry Birds game instance (game level) requires reasoning about object movements in complex physical structures \cite{Peng2014,Walega2016}. Second, there are many game levels and level generators \cite{Stephenson2017,StephensonRenz2017, Gamage2021} that enable us to evaluate our difficulty measure on a diverse set of levels. Third, this is an ideal platform to vary different physics parameters and add the type of novelties we are investigating in this study. Moreover, Angry Birds is a very popular domain among AI researchers with several long-running competitions associated with it \cite{Renzet2015,Renzet2019}. 

\section{Overview of the Difficulty Measure Formulation} \label{Difficulty Formulation}

In this section, we present a high-level overview of our difficulty measure formulation for novelty detection in physical domains. 
First, we define the following terms to aid our explanations. 

\begin{figure}[t]
\centering
\includegraphics[width=0.85\columnwidth]{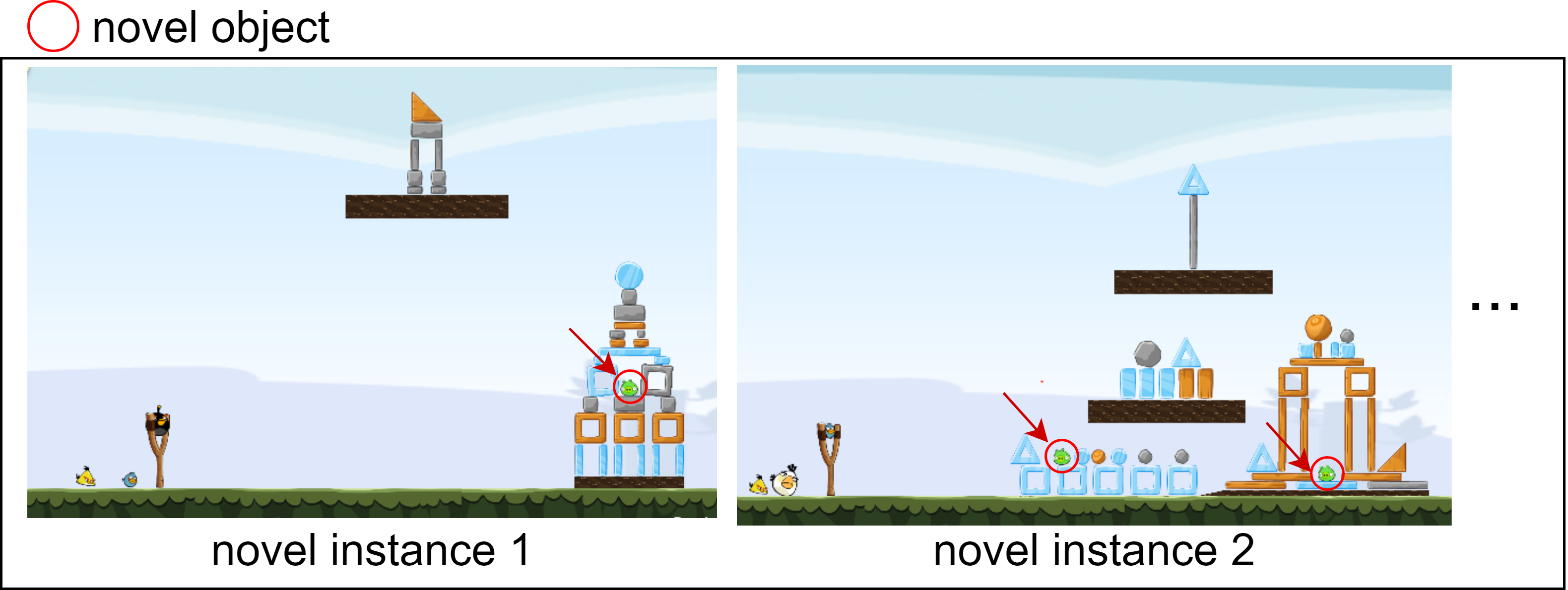} 
\caption{The figure shows a set of novel instances. Each instance contains one or more novel pigs denoted by the red circle and a set of normal objects. Note that, in this paper, we focus on novel objects with the same appearance as non-novel objects but with different physical parameters. Therefore, novel objects cannot be distinguished visually even though their physical parameters have changed.}
\label{Fig instance_environment}
\end{figure}

\begin{itemize}
    \item Each object consists of a set of appearance-related parameters and a set of physical parameters. There is a predefined many-to-one mapping from appearance parameters to physical parameters (objects with the same appearance always have the same physical parameters and two or more objects with different appearances can have the same physical parameters).
    Objects with the same appearance are referred to as an object type. The number of object types is predefined. 
    
    \item {\it normal object}: An object that preserves the predefined mapping between appearance and physical parameters. 

    \item {\it novel object}: An object that violates the predefined mapping between appearance and physical parameters. 
    
    \item {\it normal instance}: An arrangement with a collection of {\it normal objects}.
    
    \item {\it novel instance}: An arrangement with a collection of {\it normal objects} and {\it novel objects}. (See Figure \ref{Fig instance_environment})
    
    
\end{itemize}

Our difficulty measure has three properties:
\begin{enumerate}
    \item Our difficulty measure is instance-based, i.e., we provide the difficulty of detecting novelty for a specific novel instance.
    
    \item Our difficulty measure quantifies the difficulty of detecting novelty when an agent encounters the novel instance for the first time (the agent does not attempt the instance multiple times).
    
    \item Our measure is agent independent (i.e., we do not collect data from agents to develop the measure). 
\end{enumerate}

Given below are three assumptions we make about our difficulty measure.   
\begin{enumerate}
    \item As designers of the difficulty measure, we have full understanding of the novel instance (i.e., the novel object, the location of the novel object, the changed parameters, and the value of the parameters). 
    
    \item The agent has a full understanding of the object dynamics in the normal environment.
    The agent is fully aware of how objects move without novelty and the agent can detect that the environment is novel if a change in movements happens in the novel environment (perfect novelty detection).
    
    We made this assumption because the detection difficulty can be different across agents; therefore, our measure is based on the lower bound of the detection difficulty by assuming perfect detection.
    
    \item The agent attempts to detect novelty using a sequence of interactions.
    
    This means that the agent cannot have multiple interactions at the same time. For example, in the billiards game, an agent can move only one ball at a time and in Angry Birds, an agent shoots the given birds in a sequence.
    
\end{enumerate}

\begin{figure*}[t]
\centering
\includegraphics[width=1.05\columnwidth]{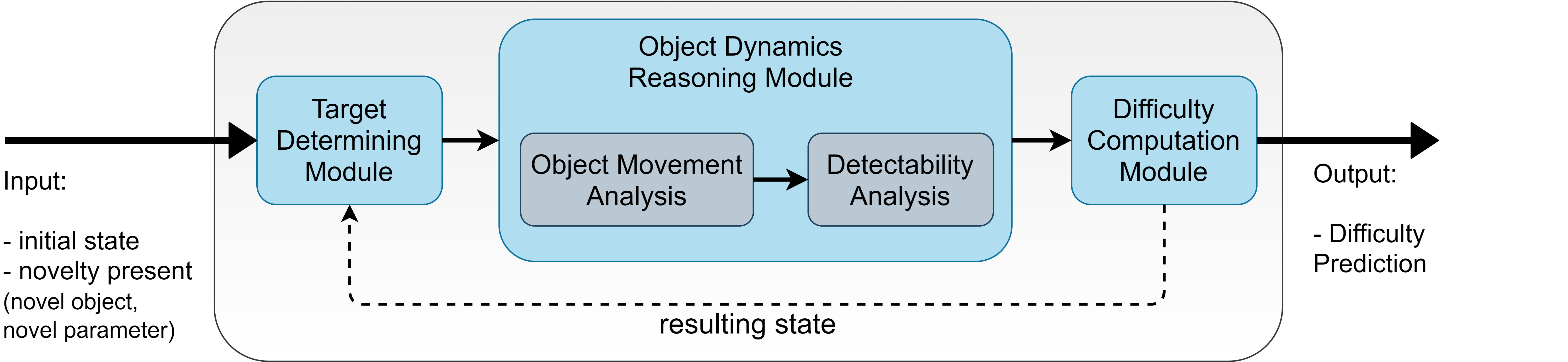} 
\caption{Overview of the method to compute the difficulty of novelty detection.}
\label{Fig difficulty framework}
\end{figure*}

Figure \ref{Fig difficulty framework} shows the main components of our difficulty measure formulation: {\it Target Determining Module, Object Dynamics Reasoning Module}, and {\it Difficulty Computation Module}. There are two inputs, the initial state of an instance (i.e., the state of an instance before any interactions) and the novelty present in the instance. Novelty present can be a set of objects with their changed physical parameter (e.g. \{(wood objects, mass), (stone objects, friction)\}).  

Our first module, the {\it Target Determining Module} takes the above two inputs and searches possible target objects, which are the objects that an agent can interact with. This module outputs all possible target objects in the given state. 

The second module, {\it Object Dynamics Reasoning Module} has two components, the {\it object movement analysis} component and the {\it detectability analysis} component. The {\it object movement analysis} component takes each target object and identifies other objects that are moved due to the interaction with the target object. Next, the {\it detectability analysis} component determines if the interaction has caused the novel object to move in a detectable way. For example, when a novel object has a different sliding friction value, an interaction that causes the novel object to fall from above would not make the novel object detectable. In contrast, an interaction that causes the novel object to slide on a surface would make the novel object detectable. 

Knowing the target objects that make detectable movements, the {\it Difficulty Computation Module} quantifies the difficulty of novelty detection to the given initial state. 
If the algorithms in the difficulty computation module require the next state to predict the difficulty, the state updates (i.e., the state after an interaction) are sent to the {\it Target Determining Module} (shown by the dotted arrow in Figure \ref{Fig difficulty framework}) and the process iterates until the detection difficulty for the instance can be calculated.  


\section{Difficulty Measure Applied to Angry Birds} \label{Difficulty Measure Formulation in Angry Birds}

This section presents each component of Figure \ref{Fig difficulty framework} in detail by considering the domain of Angry Birds. Novelties in Angry Birds can appear in any game object. When explaining our difficulty measure formulation, we do not consider the novelties that appear in the birds’ game object, as such novelties can usually be identified directly after a single shot by observing birds' behaviour. 

\begin{figure}[t]
\centering
\includegraphics[width=0.7\columnwidth]{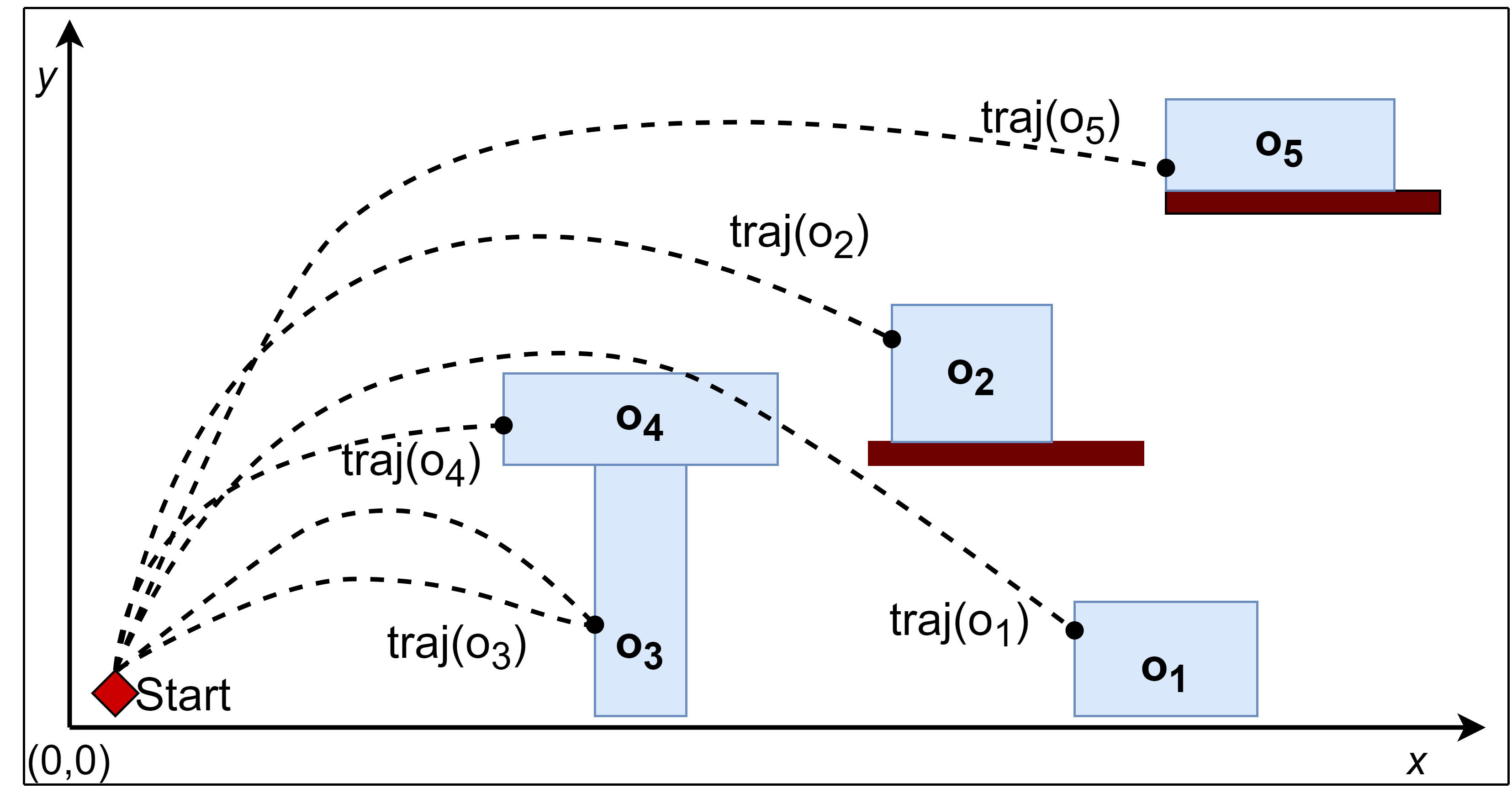} 
\caption{Representation of the object space. {\it $o_2$, $o_3$, $o_4$} and {\it $o_5$} satisfy the {\it left-of} relation to {\it $o_1$}. The trajectories to each object are denoted by the dotted line. A dot in the line represents a pixel point {\it $p_i$} $\in$ {\it P}. {\it $o_2$, $o_3$, $o_4$,} and {\it $o_5$} satisfy the {\it target} predicate. {\it $o_1$} is not a target as the {\it traj($o_1$)} intersects with {\it $o_4$}, which is in left to $o_1$. {\it $o_3$} supports {\it $o_4$}. Therefore, if {\it $o_3$} moves, {\it $o_4$} also moves: Thus, {\it impacted($o_3$,$o_4$)} is true.}
\label{Fig trajectories and targets}
\end{figure}

The first input is the initial state of the instance without any interaction. In our example domain, this is the game instance before shooting any birds. To represent the game scene, we use a 2D coordinate system where the {\it x}-axis is horizontal and oriented to the right while the {\it y}-axis is vertical and oriented to the top (Figure \ref{Fig trajectories and targets}). {\it P} denotes all pixel points in a scene. For a pixel $p_i$, {\it x($p_i$)} and {\it y($p_i$)} denote its {\it x} and {\it y} coordinates. {\it O} represents all objects in the environment. Each object $o_j$ (s.t. $o_j$ $\in $ {\it O}) comprises of a set of pixels which can be mapped to a specific object (e.g. square wood block).

The second input is the novelty present in the instance. In our example domain, novelties may appear in different object categories (i.e., wood, ice, stone, pigs), and the novel property could be any physics parameter (e.g. mass, friction, bounciness, etc). Thus, an example of the input is {\it(stone blocks, mass)}.
These inputs are sent to the {\it target determining module} to search for possible target objects. 

\subsection{Target Determining Module}

This module is used to identify the target objects. We consider the target objects are the objects that are directly reachable to interact with. We do not consider platforms as target objects as they are static. We use the following predicates to determine the targets in our example domain.

\begin{itemize}
    \item {\it left-of ($o_i$, $o_j$)}: if object $o_j$ is in left of object $o_i$ (See Figure \ref{Fig trajectories and targets}). 
    
    {\it left-of($o_i$, $o_j$)} $\equiv$ $o_i \neq o_j \land x_{max}(o_i) > x_{min}(o_j)$,
    
    where: 
    
    $x_{max}(o_i)$ and $x_{min}(o_j)$ are the maximum pixel coordinate of object $o_i$ in x direction and minimum pixel coordinate of $o_j$ in x direction respectively. 
    
    $x_{max}(o_i)$ = {\it max(x($p_k$) $\forall$ $p_k\in o_i$)},
    
    $x_{min}(o_j)$ = {\it min(x($p_k$) $\forall$ $p_k\in o_j$)}
    
    
    \item {\it traj($o_i$)}: trajectory from a starting point to object {\it $o_i$}. 
    
    As shown in Figure \ref{Fig trajectories and targets} for object $o_3$, a trajectory may contain multiple connections starting from a fixed position (slingshot in Angry Birds) to the connection point of the object. The connections can be represented using a set of points denoted by the dotted lines in Figure \ref{Fig trajectories and targets}. 
    
    We define:
    
    {\it traj($o_i$)} = $\{t_1^i, t_2^i,...,t_n^i\}$
    
    where,
    $t_k^i = \{p_{1k}, p_{2k},...,p_{nk}\}$
    
    i.e., a set of points that belong to one of the parabola trajectories and only $p_{n\cdot}\in o_i$.
    
    
    
    
    \item {\it target($o_i$)}: if object {\it $o_i$} is a target object. 
    
    {\it $o_i$} is a target if the object is directly reachable, i.e., there is at least one trajectory to {\it $o_i$} such that trajectory points do not intersect with any object with {\it left-of} relation to {\it $o_i$} according to our domain.
    
    {\it target$(o_i)$} $\equiv (\exists$ $t^i$ $\in$ $traj(o_i))$ $\land$ $t^i$ $\notin$ $o_j$ $\forall$ $\{ o_j:$ $left-of(o_i,o_j)$ $\forall o_j \in O \}$ 
    

\end{itemize}

Similar to the above {\it left-of} relation, we can define {\it right-of, below,} or {\it above} relations according to the requirement in each domain. We can also define {\it traj($o_i$)} and {\it target($o_i$)} specific to each domain. For example, if the way to interact with the objects is to drop an object from above, {\it traj($o_i$)} should be defined according to the path taken by the object in gravitational free fall and {\it target($o_i$)} is determined by the trajectories that do not intersect with the objects in {\it above} relation to {\it target$(o_i)$} $\equiv (\exists$ $t^i$ $\in$ $traj(o_i))$ $\land$ $t^i$ $\notin$ $o_j$ $\forall$ $\{ o_j:$ $above(o_i,o_j)$ $\forall o_j \in O \}$ 



\subsection{Object Dynamics Reasoning module}

After target objects are determined, it is necessary to identify the objects that can be moved due to the interactions with the target objects. This is achieved by using the {\it object movement analysis} component. We instantiate all components using our proposed qualitative physics algorithms. If the novel objects are among the impacted objects identified (defined below) or the target objects, the {\it detectability analysis} component captures if the novel objects move in a detectable way. We first define the following to aid the explanations of the methods used in the two components.

\begin{itemize}
    \item {\it novel-object($o_i$)}: if object {\it $o_i$} is a novel object. As defined earlier, {\it $o_i$} is a novel object if it violates the predefined mapping between appearance and physical parameters. i.e., object has changed physical parameter values. 
    \item {\it impacted($o_i,o_j$)}: if $o_j$ is moved due to the interaction of a bird with the target object $o_i$. For example, if $o_i$ supports $o_j$ and $o_i$ is hit by a bird, $o_j$ also moves (See $o_3$ and $o_4$ in Figure \ref{Fig trajectories and targets}).
    
    The reasoning for the identification of such objects is presented in the section {\it object movement analysis} (\ref{Section Object Movement Analysis}).
    
    \item {\it detectable($o_i,o_j$)}: if $o_j$ moves in a detectable way due to the interaction of a bird with the target object $o_i$. detectable($o_i,o_j$) returns true when $o_j$ is a novel object and {\it impacted($o_i,o_j$)} is true and $o_j$ is impacted by the target object $o_i$ in a detectable way. 
    
    A case-based exploration on the detectability of the object movements is conducted in section {\it detectability analysis} (\ref{Section Detectability Analysis}).
\end{itemize}

\subsubsection{Object Movement Analysis} \label{Section Object Movement Analysis}

\begin{figure*}[t]
\centering
\includegraphics[width=1.0\columnwidth]{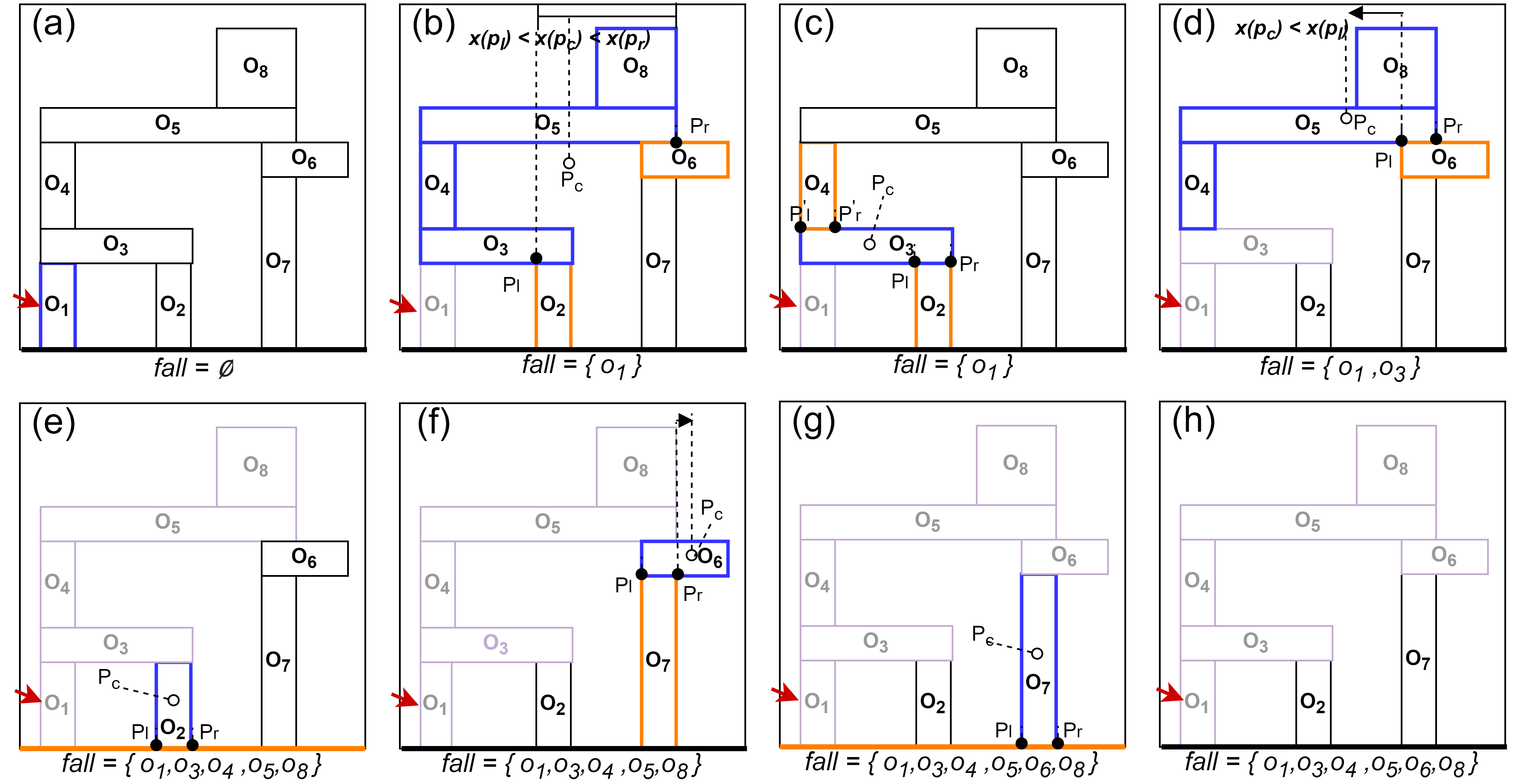} 
\caption{An illustration of the {\it vertical impact} algorithm. The algorithm comprises of 8 steps. The red arrow indicates the target object. The blue-lined blocks are the objects that are processed by the algorithm at each step. Objects in orange lie directly under (or directly above) the objects in blue. At each step, the algorithm creates substructures and reasons on the stability based on the left-most ($p_l$) and right-most ($p_r$) connection points and updates the {\it fall list}. $p_c$ is the centre of mass of the substructure under consideration. At the end of the 8 steps, the algorithm returns the list of objects that may fall after the interaction with a target object. In the figure, {\it vertical impact($o_1$) = \{$o_1,o_3,o_4,o_5,o_6,o_8$\}}. $o_1$ target object satisfies the {\it impacted} predicate with the objects in the fall list, and {\it impacted($o_1,o_2$)} and {\it impacted($o_1,o_7$)} are false. }
\label{fig verticle impact}
\end{figure*}

This section presents the qualitative physics approach used in identifying the objects that satisfy the {\it impacted} predicate presented above. i.e., we identify which objects move after an interaction with a target object. We use two algorithms 1) based on the stability, \cite{Walega2016} 2) based on the force propagation in the horizontal direction (Algorithm \ref{alg approximate horizontal influence}).  
We used the algorithm {\it vertical impact} proposed by Walega et al.\cite{Walega2016} to reason about the stability of the objects. We also propose a new algorithm, {\it approximate horizontal influence} to check the impact on the objects located in the horizontal direction.




\textbf{\textit{Vertical impact:}} This algorithm recursively checks the objects in a structure starting from the object that is directly impacted and returns a list of objects that may fall.

It exploits the rule which is the basis for stability investigation, “an object is stable if the vertical projection of the centre of mass of the object falls into the area of support base” \cite{Peng2014}. The algorithm contains eight steps where at each step object relationships are examined and substructures are constructed. The stability of objects is examined by approximating the centre of mass of substructures and their supporters. A clear explanation of the algorithm is available in the work of Walega et al.\cite{Walega2016} and Figure \ref{fig verticle impact} diagrammatically summarizes the algorithm.

\textbf{\textit{Approximate horizontal influence:}} This algorithm examines the impact a target object can cause due to the force propagation on the objects located horizontally to the target.

We start by analysing if the target object can get destroyed due to the interaction. If it is not destroyed, we check if the object will slide or it will flip as a result of the interaction (collision). Destruction of the target object heavily depends on the materials and the types of the two colliding objects and the velocity at the collision. In our example domain, we define the following predicate by considering the object materials (e.g., wood, ice, stone, pig) and the bird (e.g., red, blue, yellow). We approximate the velocity at the collision by using the law of energy conservation.

{\it object-destroy($o_i$) $\equiv$ $o_i^{life}$ – damage $<$ 0}. $o_i^{life}$ is the object life and it depends on the material of the object and type of it (e.g. square wood-block, rectangular ice-block). This is a constant value for a specific object. {\it damage} depends on the type of the bird used and the relative velocity at collision. Damage caused by a bird type is a fixed value for a specific object, $o_i^{bird\_damage}$. {\it damage} can be approximated as $o_i^{bird\_damage}$ $\times$ {\it relative-velocity} at collision.  {\it relative-velocity} can be approximated using the law of energy conservation. Thus, the final formulae for the {\it object-destroy($o_i$)} predicate can be given as follows:


{\it object-destroy($o_i$)} $\equiv$ $\big($ $o_i^{life}$ $-$ $o_i^{bird\_damage}$ $\times$ $\sqrt{{\it k1 \times (y_{start} - y_{target}) + k2_{bird}}}$   $\big)$ {\it $<$ 0},\\
\noindent
where, $k1$ is an experimentally fixed constant value, and  $k2_{bird}$ is a value based on the initial kinetic energy of the bird (In Angry Birds, the value only depends on the bird mass as the initial launch velocity is fixed because agents use the slingshot with full stretch). $(y_{start}- y_{target})$ gives the difference in height between the slingshot and the target object.

If the {\it object-destroy($o_i$)} predicate is false, we check the {\it object-flip($o_i$)} predicate by considering object dimensions.

{\it object-flip($o_i$)} $\equiv$ $\frac{y_{max}(o_i) - y_{min}(o_i)}{x_{max}(o_i) - x_{min}(o_i)}$ $>$ {\it $k_{flip}$},

where: 

$y_{max}(o_i)$ and $y_{min}(o_i)$ are the maximum pixel coordinate of object $o_i$ in y direction and minimum pixel coordinate of $o_j$ in y direction respectively. The $k_{flip}$ is an experimentally fixed constant value.

{\it $k_{flip}$= flipping threshold},

$y_{max}(o_i)$ = {\it max (y($p_j$) $\forall$ $p_j\in o_i$)},

$y_{min}(o_i)$ = {\it min (y($p_j$) $\forall$ $p_j\in o_i$)},

and $x_{max}(o_i)$, $x_{min}(o_i)$ are as defined previously.

\begin{figure}[t]
\centering
\begin{subfigure}{.43\textwidth}
  \centering
  \includegraphics[width=0.98\linewidth]{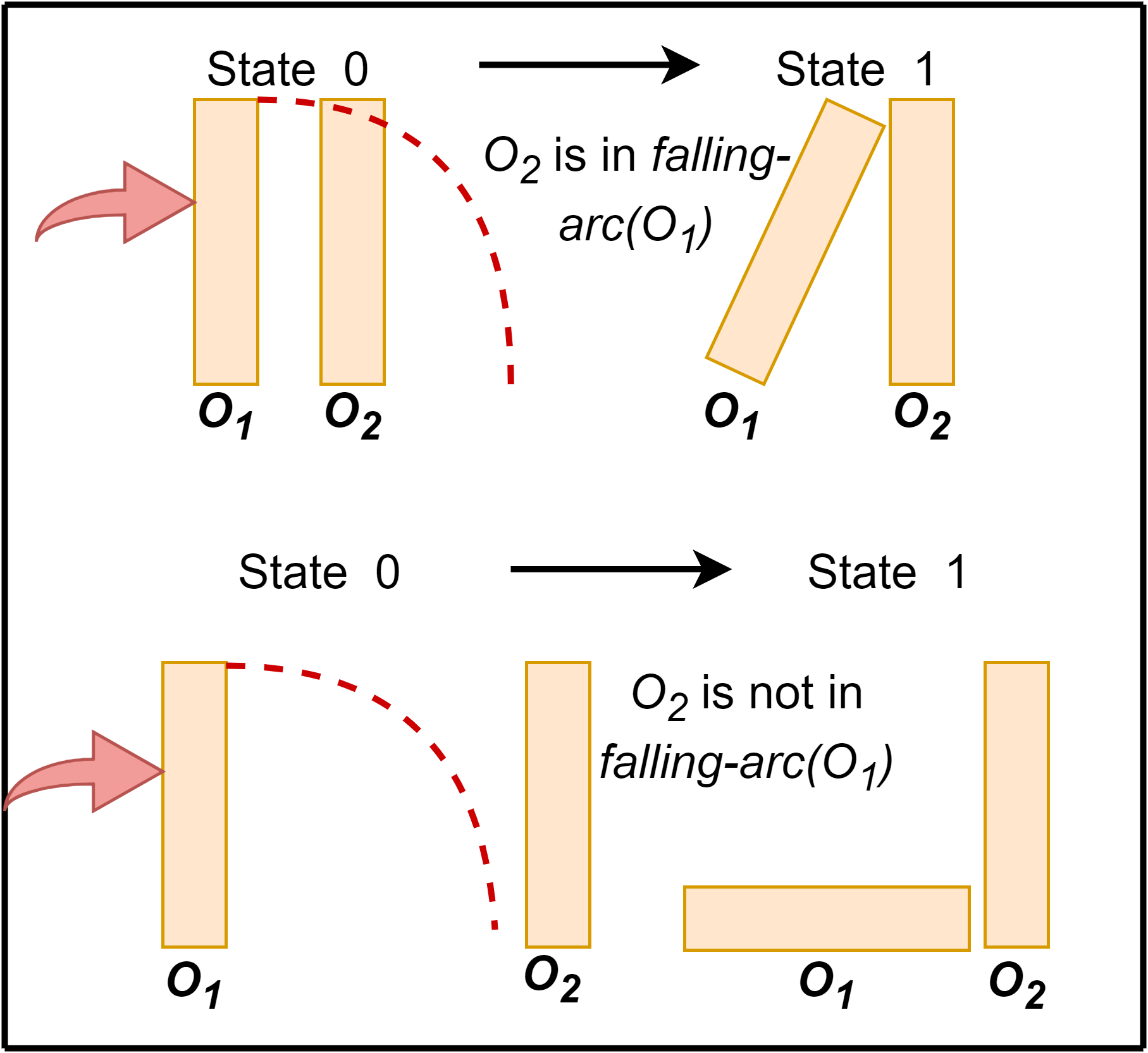}  
  \caption{}
  \label{fig:sub-first2}
\end{subfigure}
\begin{subfigure}{.43\textwidth}
  \centering
  \includegraphics[width=0.98\linewidth]{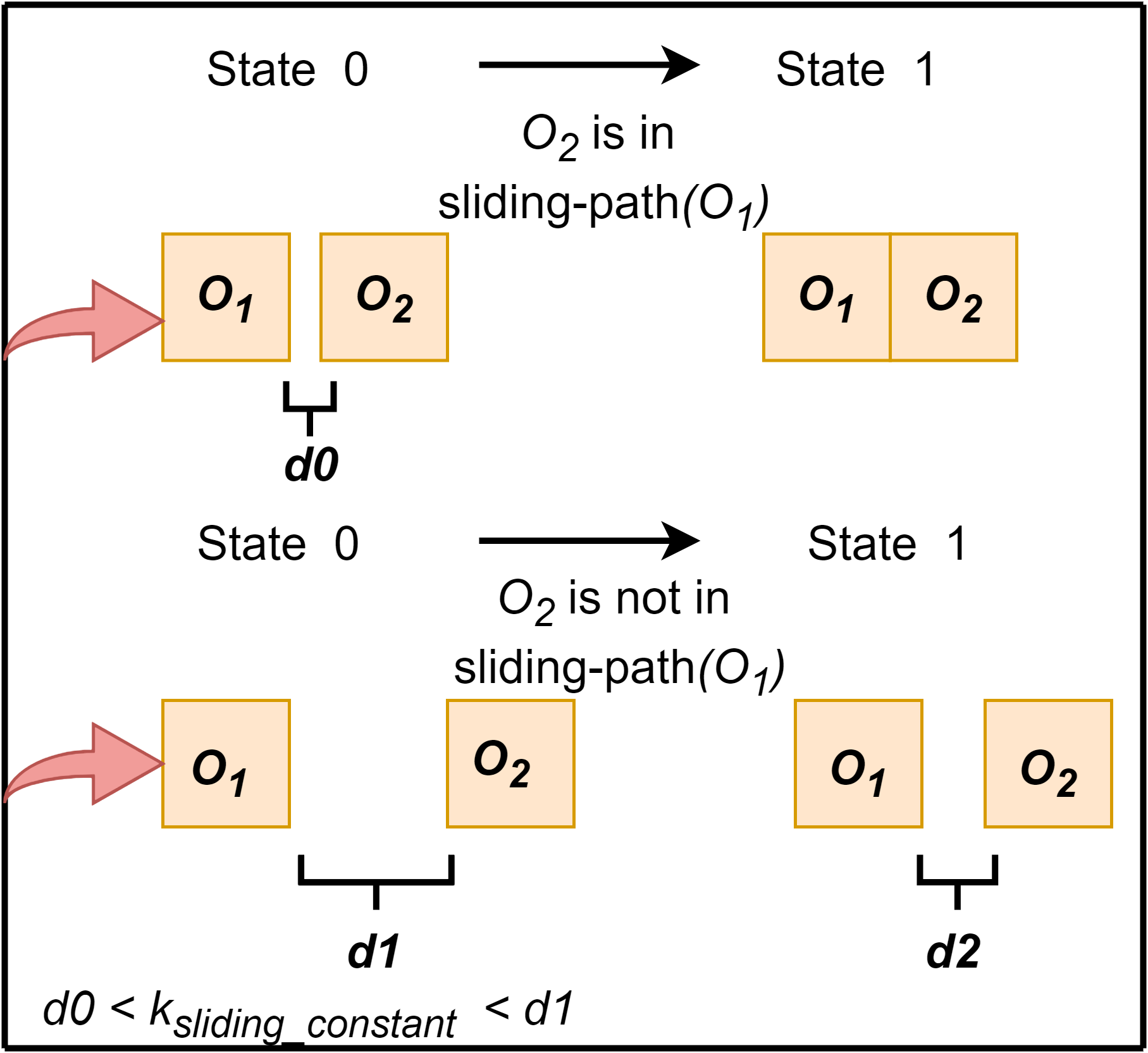}  
  \caption{}
  \label{fig:sub-second2}
\end{subfigure}
\caption{ (a) shows examples for {\it falling-arc($o_1$)} and (b) shows examples for {\it sliding-path($o_1$)}}
\label{Fig falling and sliding arc}
\end{figure}

These predicates hold the basis for the {\it approximate horizontal influence} algorithm. A pseudo-code of the process is demonstrated in Algorithm \ref{alg approximate horizontal influence} and Figure \ref{Fig falling and sliding arc} explain the terms {\it falling-arc($o_i$)} and {\it sliding-path($o_i$)} used in Algorithm \ref{alg approximate horizontal influence}. 
\begin{itemize}
    \item For a circle $C$, with centre $(x_{max}(o_i),y_{min}(o_i))$ and radius $(y_{max}(o_i)-y_{min}(o_i))$, let $q1$ be the set of pixel points in the first quadrant of $C$. {\it falling-arc($o_i$)} returns the list of objects within the falling arc of object $o_i$ (See Figure \ref{fig:sub-first2}). We define {\it falling-arc($o_i$)} as follows:
    
    {\it falling-arc($o_i$)} $\equiv$ $\{ o_j \in O \mid o_j \neq o_i \land (o_j \cap q1) \forall o_j \in O\}$
    
    \item {\it sliding-path$(o_i)$} returns the list of objects within the path the object $o_i$ slides (See Figure \ref{fig:sub-second2}). We define {\it sliding-path$(o_i)$} as follows:
    
    {\it sliding-path$(o_i)$} $\equiv$ \{ $o_j \in O$ $\mid$ $o_i \neq o_j$ 

    $\land$ $(x_{max}(o_i)<x_{min}(o_j)<x_{max}(o_i)+k_{sliding\_constant})$
    
    $\land$ (($y_{min}(o_i) < y_{max}(o_j) < y_{max}(o_i)) \lor(y_{min}(o_i) < y_{min}(o_j) < y_{max}(o_i)))$ 
    
    $\forall o_j \in {\it O}$\}
    
    where, $k_{sliding\_constant}$ is an experimentally determined distance that approximates the distance an object can slide after a collision. 
\end{itemize}
In Algorithm \ref{alg approximate horizontal influence} (line 8), we only limit to a single {\it closest\_object} obtained from either the {\it falling-arc} or {\it sliding-path} according to the experimentation with our example domain. However, this can be altered according to the domain.
The output of the {\it object movement analysis} module is the list of impacted objects obtained from the {\it vertical impact} algorithm and the {\it approximate horizontal influence} algorithm.

\begin{algorithm}[tb]
\caption{Approximate horizontal influence}
\label{alg approximate horizontal influence}
\textbf{Input}: State representation of objects, target object ({\it $o_i$}) \\
\textbf{Output}: List of impacted objects 

\begin{algorithmic}[1] 
\STATE Initialize {\it horizontal-propagation(HP)\_impact\_list}
\IF {$\neg$ ({\it object-destroy($o_i$)})}
\IF {{\it object-flip($o_i$)}}
\STATE {\it pending$\_$list} = {\it falling-arc($o_i$)}
\ELSE
\STATE {\it pending$\_$list} = {\it sliding-path($o_i$)} 
\ENDIF
\STATE {\it closest$\_$object = $o_j \mid min (x_{min}(o_j)-x_{max}(o_i)$ $\forall$ $o_j \in pending\_list)$} 
\STATE Add {\it vertical impact}({\it closest-object}) to {\it HP\_impact\_list}
\ENDIF
\STATE \textbf{return} {\it HP\_impact\_list}
\end{algorithmic}
\end{algorithm}

\subsubsection{Detectability Analysis} \label{Section Detectability Analysis}

This section presents the case-based exploration in identifying the {\it detectable} predicate presented above. 
Once the set of impacted objects are available, we can categorize each object into at least one of the below cases that represent observable features in a physical world. 
The observable movement of the directly-hit object (i.e., target object) can be explained using the first three cases.

\begin{itemize}
    \item Case 1: Directly hit and destroys
    \item Case 2: Directly hit and flips
    \item Case 3: Directly hit and slides
\end{itemize}

Apart from these three special cases, all objects subject to at least one of the following six cases.

Case 4 and 5 focus on object rotation. 
We assumed that rotation of the impacted objects directly above and very close to static structures (ground or a platform) is hardly observable. Other objects could rotate due to the collisions with objects and there is a chance of observing the rotation when objects fall.


\begin{itemize}
    \item Case 4: Falls from the top without rotating
    \item Case 5: Falls from the top while rotating
\end{itemize}

Case 6 and 7 focus on the objects that slide. The object may slide and stop, or it might fall if it’s located above the ground based on the sliding path.

\begin{itemize}
    \item Case 6: Slide and stop
    \item Case 7: Slide and fall down
\end{itemize}

Case 8 and 9 focus on the objects which flip. Similar to the above two cases, it may either fall or stop based on its location.

\begin{itemize}
    \item Case 8: Flips and stop
    \item Case 9: Flips and fall down
\end{itemize}

\begin{figure*}[t]
\centering
\includegraphics[width=1.0\columnwidth]{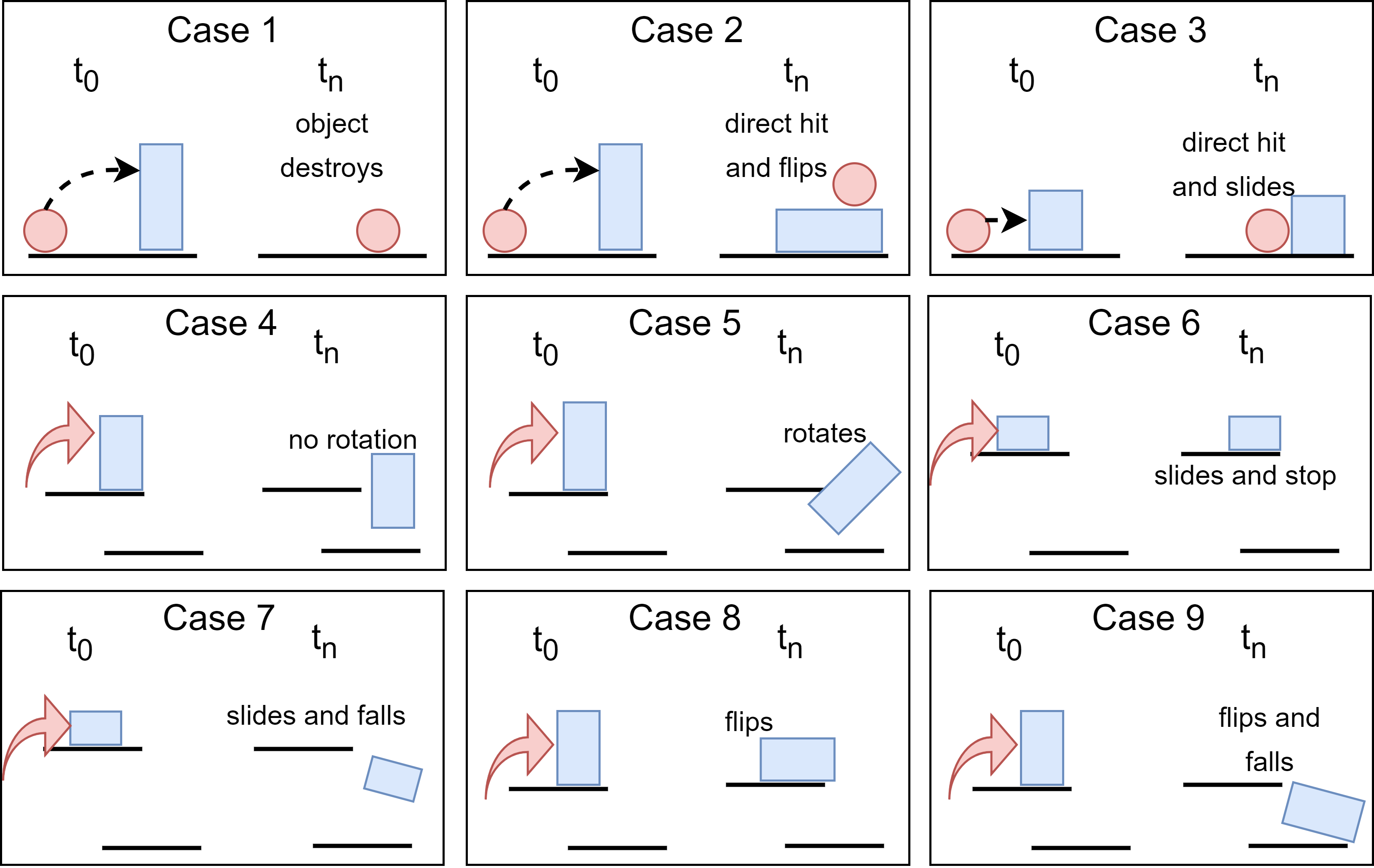} 
\caption{An illustration of the nine cases used in case-based exploration in identifying the \textit{detectable} predicate. }
\label{fig cases for detectable module}
\end{figure*}

The nine cases cover the majority of observable movements in Angry Birds (See Figure \ref{fig cases for detectable module}). However, there could be situations such as observable object rotations in objects closer to the ground that may be not captured using the nine cases. 

To evaluate if the novel object is detectable, we check if the object is moved in a detectable manner by considering the changed attribute along with the object type. Consider below two examples.

{\bf Example 1}: Novelty in “friction” of stone blocks - If at least one impacted stone block satisfies the requirements for case 3, 6, or 7, we can detect the novelty (as friction changes can be observed when the object slides).

{\bf Example 2}: Novelty in “bounciness” of wood objects - If at least one impacted wood object satisfies the requirements for case 2, 3, 4, 5, 6, 7, 8, or 9, we can detect the novelty (as bounciness can be observed when objects collide).

The output of this module enables to capture the objects that satisfy {\it detectable} predicate for each target object.

\subsection{Difficulty Computation Module}

This component quantifies the difficulty of detecting novelty for each game instance. We propose two algorithms to calculate the detection difficulty.
Factors including the novelty in the object, the placement of the objects, the number of detectable objects, the number of reachable objects, and the number of interactions available (number of birds in Angry Birds) are considered when developing both methods.

We define the following to identify the most influential target object to interact with (i.e., the target object that gives the most information about objects movements. We refer to this as the {\it best-target}).

\begin{itemize}
    \item {\it impact-score($o_i$)}: The heuristic impact score of {\it target($o_i$)} is defined based on the objects moved and the novelty introduced. 
    
    {\bf Example 1}: If the novelty is in only one object in the instance, the {\it score} per each object moved = 1
    
    {\bf Example 2}: If the novelty is in objects with the same material (wood, ice, stone), the {\it score} per material moved=1
    
    {\bf Example 3}: If the novelty is in object types and if the player is informed that the wood objects are not novel, the {\it score} per each wood object moved = 0, the {\it score} per other types of objects moved = 1
    
    
    {\it impact-score($o_i$)} = $ \sum_{o_j \in O | impacted(o_i,o_j)} score_{o_j} $

    \item {\it best-target}: The target object with the highest {\it impact-score}. If there are multiple objects with the same impact-score, the first object from all objects is selected as the best-target. 
    
	
	{\it best-target} $\equiv$ $o_i$ $\mid$ {\it target($o_i$)} $\land$ {\it impact-score($o_i$) = max(impact-score($o_i$))} $\forall$ $o_i \in {\it O}$
	
\end{itemize}

\subsubsection{Probabilistic interaction difficulty (PID)}

\begin{algorithm}[tb]
\caption{Probabilistic interaction difficulty}
\label{alg: probabilistic interaction difficulty}
\textbf{Input}: State representation of objects {\it (O)} 

\textbf{Output}: {\it PID} 

\begin{algorithmic}[1] 
\STATE Initialize {\it PID = 0}
\FOR{$i$ in {\it total\_number\_of\_interactions}}
\STATE  $N_i$ = $\mid$ \{ $o_j$ $\in O$ $\mid$ {\it target($o_j$)} $\forall$ $o_j \in {\it O}$ \} $\mid$
\STATE  $n_i$ = $\mid$ \{ $o_j \in O$ $\mid$ ({\it target($o_j$)} $\land$ $\exists$ $o_k \in O$ s.t. {\it novel-object($o_k$)} $\land$ {\it detectable($o_j,o_k$)}) $\forall$ $o_j, o_k \in {\it O}$ \} $\mid$
\STATE  $M_i$ = ($N_i$ – $n_i$) / $N_i$
\STATE {\it PID} += $M_i$
\IF{$M_i$ $\neq$ 1}
\STATE break
\ELSE
\STATE Shoot at the {\it best-target}
\STATE Update state of objects
\ENDIF
\ENDFOR
\STATE {\it PID = PID / total\_number\_of\_interactions}
\STATE \textbf{return} {\it PID}
\end{algorithmic}
\end{algorithm}

Algorithm \ref{alg: probabilistic interaction difficulty} is based on the intuition that the probability of novelty detection depends on the number of novel objects available. Intuitively, if the probability of finding a target that impacts the novel object in a detectable way is lower, the difficulty is higher. {\it PID} is initialized at zero, and the algorithm loops over the number of possible interactions (i.e., number of birds available in Angry Birds) while updating the {\it PID}. To proceed to the next interaction, it is assumed that the agent shoots at the best-target and the objects in the environment are updated along with the search space (which objects to explore next). Note the terms, $N_i$ represents the total number of target objects and $n_i$ represents the total number of target objects which makes the novel object move in a detectable way in the given state. Thus, $M_i$ is the proportion of targets that do not yield a detectable movement. At the end of the computation, {\it PID} is normalized to [0,1], where 1 indicates the highest difficulty and {\it PID} is unit-less. One limitation of this algorithm is that it only considers the {\it best-target} when updating the next state instead of considering all possible targets. This is due to time restrictions, and works under the assumption that an intelligent agent would always select the {\it best-target}. 

\subsubsection{Best-shot interaction difficulty (BID)}

\begin{algorithm}[tb]
\caption{Best-shot based interaction difficulty}
\label{alg: best-Shot based interaction difficulty measure}
\textbf{Input}: State representation of objects 

\textbf{Output}: {\it BID} 

\begin{algorithmic}[1] 
\STATE Initialize {\it BID = 0}
\STATE Initialize {\it detection\_flag = False}
\FOR{$i$ in {\it total\_number\_of\_interactions}}
\STATE {\it BID = BID + 1}
\IF {(\it $\exists o_j \in O \mid novel-object(o_j) \land detectable(o_k*,o_j)$)}
\STATE {\it detection\_flag = True}
\STATE break
\ELSE
\STATE Shoot at the {\it best-target}
\STATE Update state of objects
\ENDIF
\ENDFOR
\IF {{\it detection\_flag = False}}
\STATE {\it BID = {\it total\_number\_of\_interactions} + 1}
\ENDIF
\STATE {\it BID = (BID - 1) / total\_number\_of\_interactions}
\STATE \textbf{return} {\it BID}
\end{algorithmic}
\end{algorithm}

Algorithm \ref{alg: best-Shot based interaction difficulty measure} is inspired by an intelligent human-like agent and is based on the interaction which yields the most information. Here we try to maximize the chance of novelty detection by making the most influential interaction (i.e., always shooting at the {\it best-target:} $o_k$*). The algorithm loops over the number of possible interactions that can be made (i.e., number of birds available in Angry Birds): if the novelty is undetectable by shooting at the best-target, it proceeds to the next after updating the environment, the search space (which objects to explore next), and {\it BID}. Similar to Algorithm \ref{alg: probabilistic interaction difficulty}, {\it BID} is normalized to [0,1], where 1 indicates the highest difficulty and is unit-less.

These two difficulty algorithms can be used separately or collectively according to the suitability of the study. We have used the two algorithms collectively in our experimental evaluation presented in Section \ref{Experimental Evaluation}. 

\section{Experimental Evaluation} \label{Experimental Evaluation}

As the difficulty measure we propose is a general measure, we conducted an experiment to analyse the relationship between the difficulty of novelty detection when computed from the proposed measure compared to that for humans. 
Human participant experiments were approved by the Australian National University human ethics committee under the protocol 2020/717. We gathered data from 20 voluntary participants in Angry Birds. Participants represented males and females and were in the age range of 20-35. Participants did not have any prior knowledge about the tested novelties. All participants provided consent to use their play data. We first provided 10 game instances without novelty from an Angry Birds levels generator \cite{StephensonRenz2017}. This allowed the participants to become more familiar with the normal game physics and object dynamics. These participants could play the game instances any number of times in any order. Next, 15 instances with three novelty types were provided and the detection difficulty of each instance was calculated using our proposed approach in advance (See Section \ref{Game Selection}). We selected 15 instances due to time constraints, as each participant takes approximately 2-3 minutes to play a novel instance, and we selected 3 different novelties to allow for varying difficulties of detection. The participant was allowed to play the novel instance only once to detect if there is any novelty in the game objects. If the novelty was detected, we recorded the number of interactions (number of shots in Angry Birds) the participant used to detect that novelty. We also requested the participant to provide a simple description of the observation to validate the results. Each participant took approximately 40-50 minutes to complete the experiment. 

The novelties we evaluated were applied to all game objects with the chosen material (e.g., all wood blocks in the game have the novel property, all pigs in the game have the novel property). A novel game instance only contained a single novelty type. That is, a novelty only appears in a single object type (e.g., A single game instance does not contain novel wood blocks and novel pigs). This controlled setup is used to validate our difficulty measure even though it can be applied without the given limitations. The novelties we generated are as follows:

\begin{itemize}
    \item \textbf{Type 1 (T1):} The parameter \textit{gravity scale} of pigs is decreased twice the original value. Pigs fall down slower due to this novelty.
    \item \textbf{Type 2 (T2):} The parameter \textit{bounciness} of wood objects is increased by four times the original value. This makes the wood objects bouncier.
    \item \textbf{Type 3 (T3):} The parameter \textit{life} of stone objects increased by five times. This makes stone blocks more difficult to destroy.
\end{itemize}

\subsection{Game Instance Selection} \label{Game Selection}

\begin{figure*}[t]
\centering
\includegraphics[width=1.0\columnwidth]{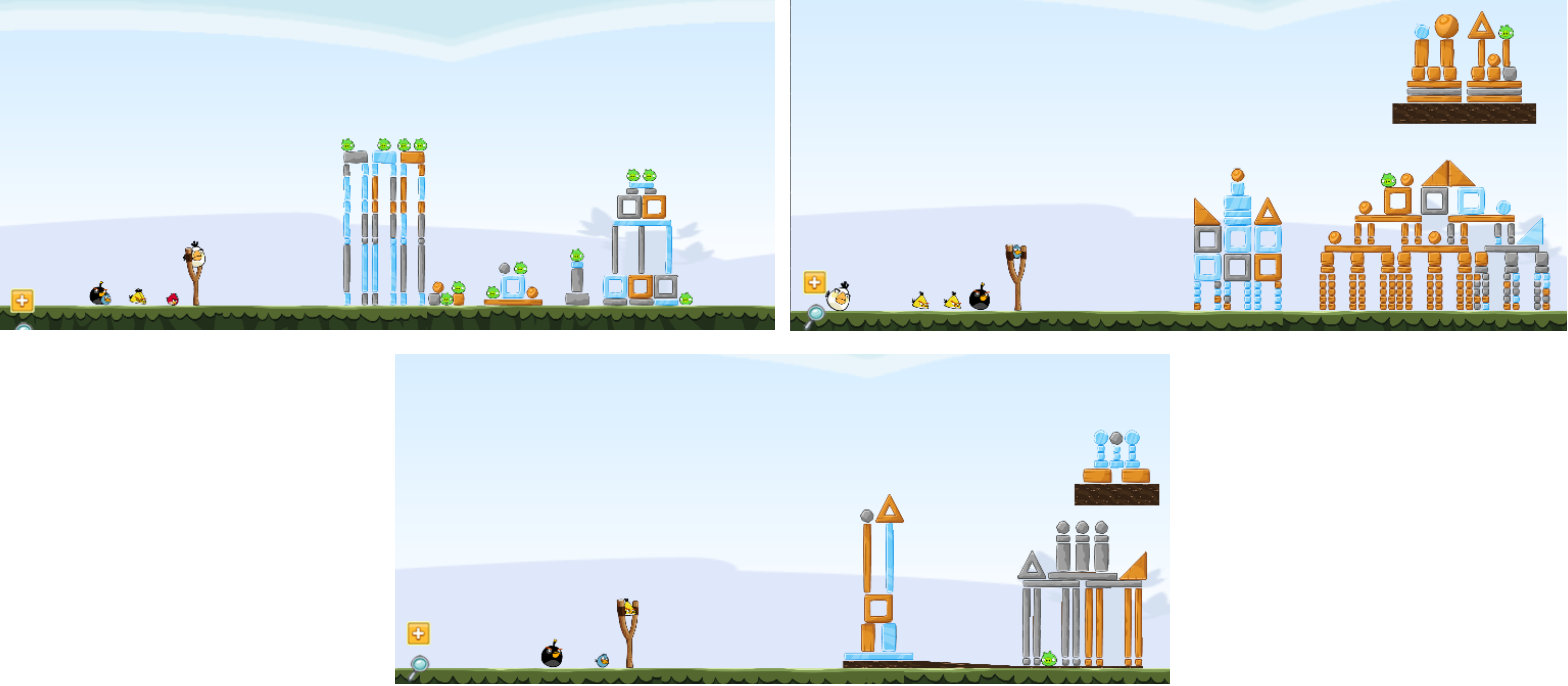} 
\caption{Example novel game instances provided for human participants. The three instances belong to the three novelty types T1, T2, and T3 respectively.  The novelties cannot be distinguished visually from the game instances without interacting.}
\label{fig example levels for humans}
\end{figure*}

A set of 100 game instances was generated from the state-of-the-art Angry Birds level generator \cite{StephensonRenz2017} and the novelty game instances were created for each novelty type. We then computed difficulty using the two algorithms, {\it PID} and {\it  BID}, for each instance. We combined the two values using: {\it Difficulty Value = $\alpha$ PID + (1 - $\alpha$) BID}, where $\alpha \in$ [0,1], which can be adjusted according to the importance of the two algorithms in an experiment. In our experiment, we considered $\alpha$ to be 0.5 to give equal importance.
Game instances within each novelty type were then classified into three categories: {\it easy, medium,} and {\it hard} based on the distribution of the difficulty values. Game instances with values less than the value at 33.33\% percentile, in between 33.33\% and 66.67\%, and values higher than 66.67\% were considered as {\it easy, medium}, and {\it hard} instances respectively. 
The game instances used for the experiment were selected randomly from each category. However, different techniques such as harmonic mean or clustering methods could also be used to categorize easy/medium/hard based on the data available. Example game instances selected for the experiment are shown in Figure \ref{fig example levels for humans}.

\subsection{Results}

\begin{figure}[t]
\centering
\begin{subfigure}{.45\textwidth}
  \centering
  \includegraphics[width=0.99\linewidth]{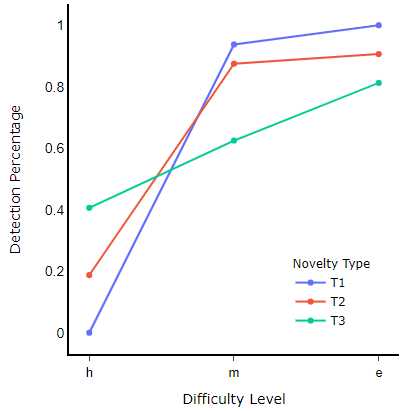}  
  \caption{}
  \label{fig:hd 1}
\end{subfigure}
\begin{subfigure}{.45\textwidth}
  \centering
  \includegraphics[width=0.99\linewidth]{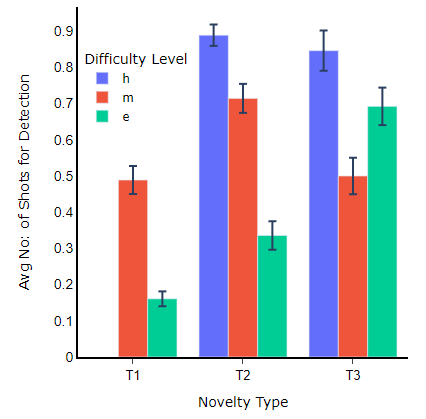} 
  \caption{}
  \label{fig:hd 2}
\end{subfigure}
\caption{Experiment results from human participants. (a) shows the percentage of novelty detection and (b) shows the average normalized number of shots for novelty detection for each difficulty level in each novelty. Error bars represent the standard error. {\it e,m,h} indicate easy, medium, and hard categories.}
\label{Fig human data}
\end{figure}

According to our difficulty measure, we expect the percentage of novelty detection to decrease in the order easy, medium, and hard (according to Algorithm \ref{alg: probabilistic interaction difficulty}). Ideally, if the novelty is detected, we expect a lower number of interactions to detect the novelty in the easy category and a higher number of interactions in the hard category (according to Algorithm \ref{alg: best-Shot based interaction difficulty measure}).

Figure \ref{fig:hd 1} illustrates the percentage of human participants who correctly detected the novelty for each novelty type in the three difficulty levels. In line with our hypothesis, the lowest percentage of detection is recorded in the hard category and the highest is recorded in the easy category. This observation is consistent for all three experimented novelty types. For the T1 novelty type, none of the participants were able to detect the novelty in the hard category, while all the participants detected it in the easy category.

Figure \ref{fig:hd 2} summarizes the average normalized number of shots needed for detection for each difficulty level for the three novelty types. That is, for each participant, the number of shots taken for detection is normalized by the total number of possible interactions (i.e., the number of birds in the game instance). For the T1 novelty type, the hard category is not presented as none of the participants detected the T1 novelty type. The medium and easy categories follow our expectation by producing a lower value for the easy category. Similarly, T2 results are also consistent with our expectation by producing the highest normalized interactions for the difficulty in the hard category and the lowest in the easy category. For T3, while the hard category gives the highest normalized interactions for detection, the medium category is lower than the easy category. According to our observations, some participants used more shots to confirm that stone-blocks have a higher health value even though they already detected this novelty earlier and some participants did not notice the change in stone-blocks at all. Overall, the difficulty of novelty detection for human participants falls in line with the calculated difficulty values from our proposed method.

\section{Application to Other Domains and Limitations}

The difficulty of the novelty detection formulation presented in Section \ref{Difficulty Formulation}, and diagrammatically shown in Figure \ref{Fig difficulty framework}, can be applied in physics environments. In Section \ref{Difficulty Measure Formulation in Angry Birds} we instantiated the components to the specific domain of Angry Birds. 
In this section, we briefly discuss how each component of our difficulty formulation can be used for another popular physical reasoning benchmark, PHYRE \cite{Bakhtin2019}. 

\begin{figure}[t]
\centering
\begin{subfigure}{.35\textwidth}
  \centering
  \includegraphics[width=0.99\linewidth]{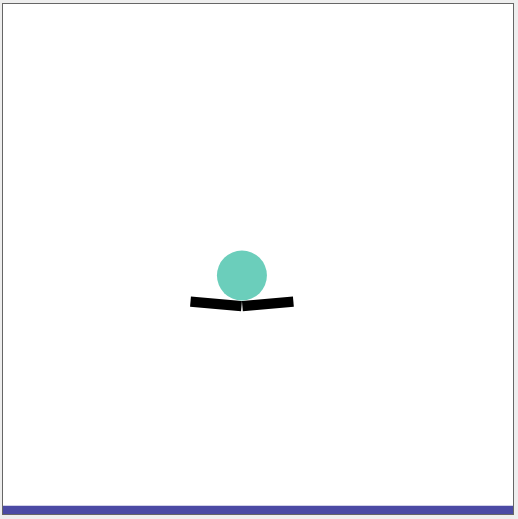}  
  \caption{}
  \label{fig:phyregen 1}
\end{subfigure}
\begin{subfigure}{.35\textwidth}
  \centering
  \includegraphics[width=0.99\linewidth]{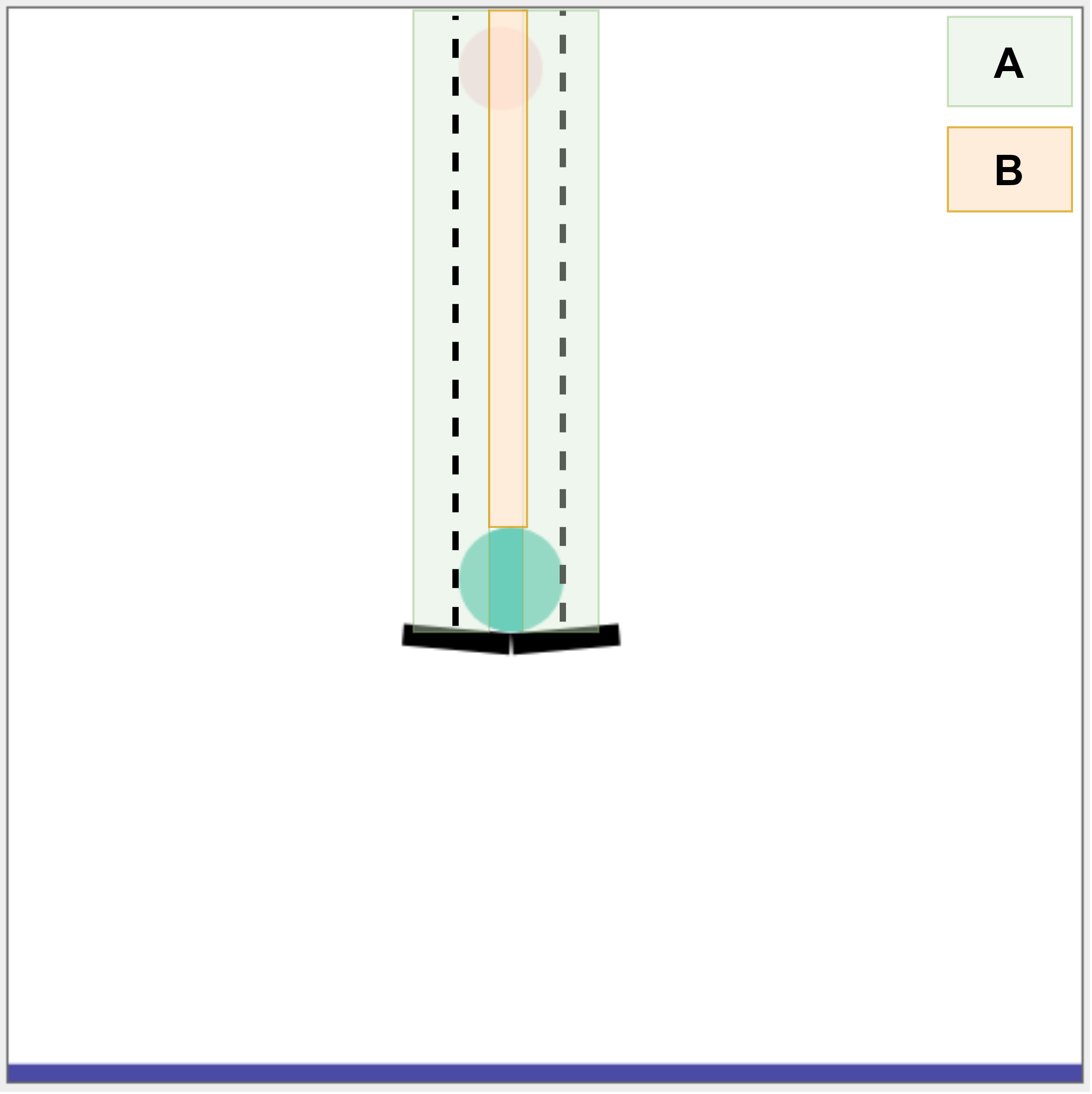}  
  \caption{}
  \label{fig:phyregen 2}
\end{subfigure}
\caption{(a) shows an example PHYRE task and (b) shows the possible regions to place the new ball such that it interacts with the green ball.
Collisions resulting by placing the center of the new ball in region A allows us to detect the bounciness change, while region B does not move the green ball in a detectable way. 
}
\label{Fig phyre generalization}
\end{figure}

\subsection{Applying to PHYRE} \label{Section Applying to PHYRE}

PHYRE is introduced as a physical reasoning benchmark, that contains a set of puzzles in a 2D physics environment. Each puzzle has a goal state and an initial state in which the goal is not satisfied. Each puzzle can be solved by placing a new body in the environment. The collisions from the new body result in the goal state if the new body is placed at the correct position. All objects in PHYRE are indestructible and non-deformable. An example task in PHYRE is shown in Figure \ref{fig:phyregen 1} at its initial state. The green ball is dynamic, the black supporter and the purple bar on the ground are static. 
The goal of this task is to make the green ball touch the purple bar by placing a new ball within the environment. 

Assume we introduce novelty to this domain and the goal of the agent is to detect the novelty. 
As an example novelty, for the task shown in Figure \ref{fig:phyregen 1}, we increase the bounciness parameter of the green ball. We assume the novelty detecting agent has prior knowledge about the expected object movements in the domain and the agent needs to place a new ball to detect novelty within the environment. 

To measure the difficulty of novelty detection in this example, we can instantiate each component of our difficulty formulation as follows: 
\begin{itemize}
    \item \textbf{Input: }As shown in Figure \ref{Fig difficulty framework}, the input contains the initial state and the novelty present. The initial state of a PHYRE puzzle is as shown in Figure \ref{fig:phyregen 1} and the novelty is present in the green ball which has an increased bounciness from the original value. 
    \item \textbf{Target Determining Module: }This module should be instantiated by formalizing the method to find target objects that can be reached by an object falling under gravity. As shown in the figure, an agent has only one target object (green ball). 
    \item \textbf{Object Movement Analysis: }This sub-module should be instantiated to identify if the movement of the target object causes other objects to move. As there are no other movable objects around the target object, the agent does not need to reason about the movement of other objects, that is, object movement analysis only considers the movement of the target object.
    \item \textbf{Detectability Analysis: }This sub-module should analyze if the novel object will move in a detectable way due to the actions of the agent. The agent can detect the change in bounciness when the green ball collides with another object (e.g., collision on the ground).  
    \item \textbf{Difficulty Computation Module: }This module should be instantiated using the outcomes of the above modules to compute the difficulty of novelty detection. Figure \ref{fig:phyregen 2} indicates possible locations to place the new object, such that it can interact with the target object (the region shown by A and B).
    Out of the possible locations, only those in region A move the green ball in a way that allows for the bounciness change to be detected. 
    Using this data, we can use the same measures formulated for Angry Birds to compute the novelty detection difficulty. In this example, 
    
    \textit{PID = area of region A/ (area of region A + area of region B)}
    
    Similarly, 
    \textit{BID = 0}, as the agent has the chance of detecting novelty from the first action. 
\end{itemize}

\subsection{Limitations of the Approach}

Even though our formulation of the difficulty of novelty detection can be applied in physical domains, each component should be instantiated to suit the domain accordingly. Our presented qualitative reasoning algorithms in {\it object movement analysis} are specific to Angry Birds, only being capable of predicting the impact of a target object's movement on other objects which are connected or located to the right side of it (as forces are applied from left to right in Angry Birds). However, our reasoning methods in Angry Birds are not capable of predicting the position where an object gets thrown to, or the possible consequences on other objects around the region it gets thrown to. We plan to extend our work with appropriate heuristics to address this in the future. 

Our formulation of the {\it detectability analysis} component does not make distinctions with regard to the possible value variations of the selected physics parameters. 
For example, if a previously empty box is now filled with 2 kg of goods or if the box is filled with 3 kg, the difference in the difficulty of detection between these two cases is not considered within this study. 

Our difficulty formulation is based on the detectable changes in object movement.
However, a non-physics related real-world novelty, such as a firmware update for a household electronic device, does not necessarily cause a change in object motion would not be captured by our proposed difficulty measure. This type of novelty requires a more advanced difficulty measure that captures the difficulty of detection caused by other factors in addition to object movement.

\section{Discussion and Conclusion}
Detecting novelty is an important capability for an intelligent system in an open-world environment. In real world situations, an agent needs to reason about physics in order to detect novel objects with different physical parameters. 
These novelties often vary in their difficulty of detection, something that has not been previously studied before this paper. However, understanding this difficulty can be an important aspect of conducting a robust and fair evaluation. Thus, we have proposed a method to quantify the difficulty of novelty detection using qualitative physics. Our method is agent independent and can be used to make more accurate conclusions about the detection capabilities of different agents. This measure was applied in the Angry Birds domain, and validated by comparing the results of the proposed measure with the performance of human participants across three different novelty types.

The different components and algorithms that were introduced in this paper can also be applied to other research problems. When formulating our novelty difficulty measure, we proposed the algorithm {\it approximate horizontal influence} that could also be used as a component for agents to predict the influence of moving an object in physical domains. This is an improvement to the prior work \cite{Peng2014,Walega2016} as it considers objects that are disconnected in the horizontal direction. 
Our difficulty formulation can also be used to create novel game instances at a predefined difficulty of novelty detection. Our difficulty formulation can be used as a component in the state-of-the-art novelty generation framework for Angry Birds \cite{Gamage2021} to generate novel game instances with a predefined difficulty. 
This facilitates research in open-world learning agent development by creating different instances with different levels of difficulty. 

We plan to extend our study to address the limitations we discussed above, to suit a wider variety of novelties and to be applicable to a wider range of domains. In this paper, we laid a foundation for quantifying the difficulty of novelty detection that helps to conduct a sound open-world agent evaluation. 

 
\section*{Acknowledgments}

This research was sponsored by the Defense Advanced Research Projects Agency (DARPA) and the Army Research Office (ARO) and was accomplished under Cooperative Agreement Number W911NF-20-2-0002. The views and conclusions contained in this document are those of the authors and should not be interpreted as representing the official policies, either expressed or implied, of the DARPA or ARO, or the U.S. Government. The U.S. Government is authorized to reproduce and distribute reprints for Government purposes notwithstanding any copyright notation herein.

\bibliography{mybibfile}

\end{document}